# On Unsupervised Training of Link Grammar Based Language Models

Nikolay Mikhaylovskiy [1, 2][0000-0001-5660-0601]

[1] Higher IT School, Tomsk State University, Tomsk, Russia, 634050
[2] NTR Labs, Moscow, Russia, 129594
`nickm@ntr.ai`

**Abstract.** In this short note we explore what is needed for unsupervised training of graph language models based on link grammars. First, we introduce the termination tags formalism required to build a language model based on a link grammar formalism of Sleator and Temperley [21] and discuss the influence of context on the unsupervised learning of link grammars. Second, we propose a statistical link grammar formalism, allowing for statistical language generation. Third, based on the above formalism, we show that the classical dissertation of Yuret [25] on discovery of linguistic relations using lexical attraction ignores contextual properties of the language, and thus the approach to unsupervised language learning relying just on bigrams is flawed. This correlates well with the unimpressive results in unsupervised training of graph language models based on bigram approach of Yuret.

## 1      Motivation

While not long ago language models were just models that assign probabilities to sequences of words [11], now they are the cornerstone of any task in computational linguistics through few-shot learning [4], prompt engineering [14] or fine-tuning [6].

On the other hand, current language models fail to catch long-range dependencies in the text consistently. For example, text generation with maximum likelihood target leads to rapid text degeneration, and consistent long text generation requires probabilistic sampling and other tricks [9]. Large language models such as GPT-3 [4] move the boundary of "long text" (far away), but do not remove the problem.

One of the sources of the above phenomenon lies in the fact that in the correlations in natural language texts decrease according to the power law of distance between the tokens. This, in turn, is considered an outcome of the hierarchical structure of human texts [1, 2]. Further, the mutual information between two tokens decays exponentially with distance between them in any probabilistic regular grammar and Markov chains, but can decay like a power law for a context-free grammar [14].

Recent results [6] indicate that, while, theoretically, RNNs are Turing complete [20], in practice, RNNs and Transformers trained with gradient descent fail to generalize on nonregular tasks, LSTMs can solve regular and counter-language tasks, and only networks augmented with structured memory can successfully generalize on context-free and context-sensitive tasks. Thus, building language models that exhibit at least hierarchical, context-free grammar-ish, slow-correlation-decay behavior may be beneficial for a variety of downstream tasks. This may be not enough to model long texts successfully because natural languages cannot be described by a context-free grammar [19], but may be a meaningful step.

## 2   Link Grammars

Grammar of a natural language is its set of structural constraints on speakers' or writers' composition of clauses, phrases, and words ([24]). The idea of basing a grammar on constituent structure dates back to Chomsky [4] and Backus [1] (see also Jurafsky and Martin [11]). Dependency grammar formalism was simultaneously proposed by Tesniere [22]. Link grammars [21] are a type of dependency grammar; these, in turn, can be converted to and from phrase-structure grammars using relatively simple rules and algorithms.

Following Vepstas and Goertzel [23], let's consider basics of Link Grammars. In a Link Grammar, each word is associated with a set of 'connector disjuncts', each connector disjunct controlling the possible linkages that the word may take part in. A disjunct can be thought of as a jigsaw puzzle-piece; valid syntactic word orders are those for which the puzzle-pieces can be validly connected. A single connector can be thought of as a single tab on a puzzle-piece (shown in **Fig. 1**). Connectors are thus 'types' X with a + or - sign indicating that they connect to the left or right. For example, a typical verb disjunct might be S −& O+ indicating that a subject (a noun) is expected on the left, and an object (also a noun) is expected on the right.

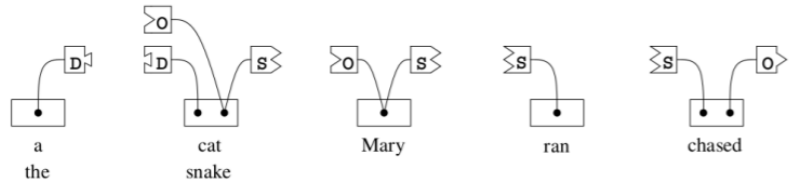

**Fig. 1.** - Link Grammar Connectors

The lexical entries in a lexicon for the above would be

```
a the: D+;
cat snake: D- & (S+ or O-);
Mary: O- or S+;
ran: S-;
chased S- & O+;
```

**Fig. 2.** - A lexicon of a link grammar

Note that though the symbols '&' and 'or' are used to write down the disjuncts, these are not Boolean operators and they do not form a Boolean algebra. They do form a non-symmetric compact closed monoidal algebra.

Vepstas and Goertzel [23] suggested representing a phrase in this notation as depicted on **Fig. 3**. In the next section, we will slightly modify this notation so that it has a more closed form.

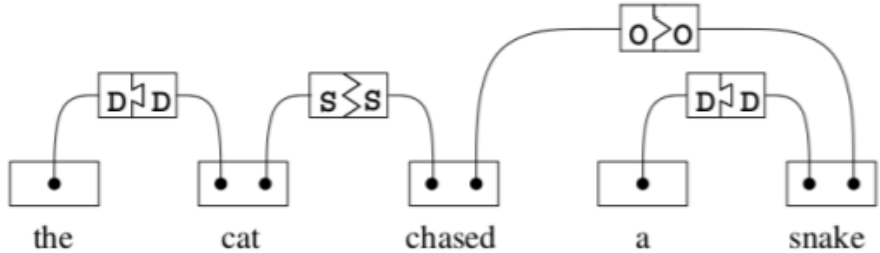

**Fig. 3.** - A phrase in a graphical notation of the Link Grammar

The above can also be written in a textual form:

$$The + D - cat + S - chased + O - snake - D + a \qquad (1)$$

One cannot guarantee a textual form for every Link Grammar graph, but we will use this notation when suitable.

## 3    Termination Tags

To make the above notation more logical, and, as we will see further, properly introduce the frequentist statistics, we introduce Termination Tags (TT) for each of the links in the link grammar, instead of original formalism that only includes LEFT-WALL. These tags are pseudo-terms that terminate the links that do not have a matching pair. Indeed, if we look at the **Fig. 3**, we will note that actually two connectors hang loose: O- connector from the 'cat' and S+ connector from the 'snake'. We depict this in **Fig. 4**:

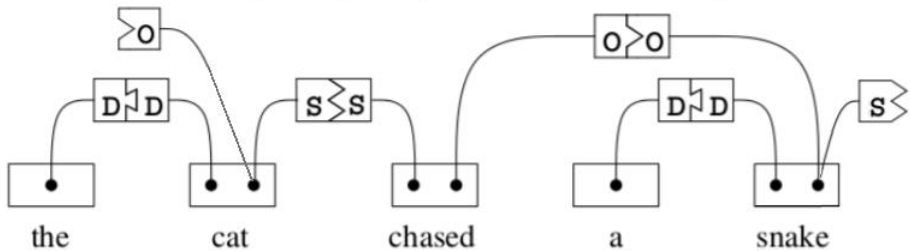

**Fig. 4.** – Loose connectors in a Link Grammar of a sentence.

Thus, we can add a terminator tag of S- type to indicate that the S+ link from 'snake' does not have a pair, and do the same for the O- link of 'cat' (see **Fig. 5**):

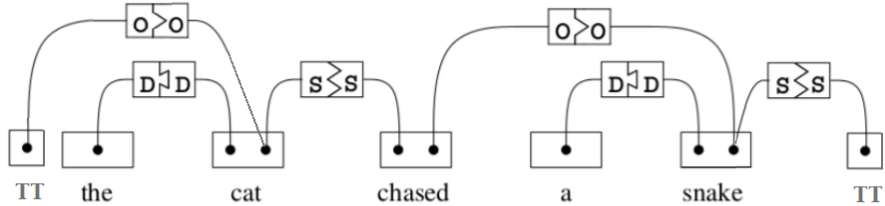

**Fig. 5.** - Terminator tag closes the link

This notation differs from the typical Link Grammar notation with beginning of sentence, but, as we will see in the next section, allows for frequentist construction of Link Grammar based language models.

## 4   Language Models Based on Link Grammars

Jurafsky and Martin [11] define language models or LMs as the models that assign probabilities to sequences of words. To achieve the goal of this work we need to extend the definition and build a model that assigns probabilities to sentences as graph structures. Probabilistic language model frameworks were created for other types of grammars equivalent to Link Grammars of [21], including [10, 13, 15]. Our goal is to add to [21] formalisms allowing language model creation.

### 4.1   Text Generation with a Statistical Link Grammar

Let's first consider the problem of text generation using a Link Grammar. Suppose we have a lexicon $\mathcal{L}$ of terms $t_k$ with their respective disjuncts, and for every connector in such a disjunct we have probabilities of words that would plug into this connector, including TT. This differs significantly from a deterministic approach of Ramesh and Kolonin [16, 17] that basically builds a surface realization.

Now we can start with any term with its disjunct, and assume that the probability of the term plugged into the disjunct depends only on the original term and the connector. In the example above, we can start with the word 'cat'. We can suppose that in the lexicon its D- connector has potential links to two terms: 'the' with probability 0.6 and 'a' with probability 0.4. Let's assume that the random sampling have returned 'the'.

The O- connector of the term 'cat' in the lexicon has potential links to three terms: 'chased' with probability 0.3, 'ran' with probability 0.2, and TT with probability 0.5. Let's assume that the random sampling have returned TT. Finally, the S+ connector of 'cat' has potential links to three terms: 'chased' with probability 0.5, 'ran' with probability 0.4, and TT with probability 0.1. Let's assume that the random sampling have returned 'chased'.

This way we have generated the closest, in a Link Grammar sense, neighbors of the starting word 'cat', as depicted on **Fig. 6**:

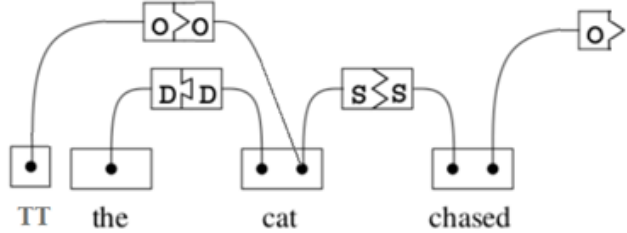

**Fig. 6.** A piece of text graph generated using the Link Grammar

We can recursively continue this procedure with each generated word that is not TT (in the example above there is only one such word – "chased") and generate a sentence graph in the Link Grammar. It is easy to see that under mild assumptions on probabilities of TT, the graph generated will almost always have a finite length. In many senses, the procedure above is similar to bigram-based text generation.

In the procedure above a termination tag may happen at both sides of a sentence, unlike the notation of [21, 23] that in a generation setting may generate infinitely.

### 4.2 Frequentist Statistics and a Link Grammar Language Model

The above can be formalized as a discrete parameterized source $S(t, l)$, where $t$ is the term of the lexicon $\mathcal{L}$ and $l$ is the specific link from the term. The source emits a connected term with a probability distribution $\{\alpha_i\}$. Each term $t_k$ that has a connector matching $l$ has a fixed probability $\alpha_k$ to be generated. $\{\alpha_i\}$ is subject to $\sum_i \alpha_i = 1$.

We can see the above probability distribution as a parametrized distribution $\{P_i(t_k, l)\}$. With a frequentist approach, we can write that the probability of term $t_i$ linked to $t_k$ with a link $l$ is $P_i(t_k, l) = \frac{C(t_k + l - t_i)}{C(t_k)}$ (remember that the expression in the numerator is a link grammar expression, not an arithmetical one). The operator $C$ counts the occurrences of its argument over a certain corpus.

If we want to go further and estimate the probability of a sentence (read: a graph in a link grammar) in a certain communicative context we can apply a chain rule of probability, taking the context into consideration. If we would be working with a sequence of words like the n-gram techniques do, we would write (Jurafsky and Martin [11]) for the probability of a sequence of $m$ words:

$$P(w_{1:m}) = P(w_1)P(w_2|w_1)P(w_3|w_{1:2})\ldots P(w_m|w_{1:m-1}) = \prod_{k=1}^{m} P(w_k|w_{1:k-1}) \quad (2)$$

and approximate it with a truncated version in a naïve Bayesian way:

$$P(w_{1:m}) = \prod_{k=1}^{m} P(w_k|w_{1:k-1}) = \prod_{k=1}^{m} P(w_{k-n:k-1}|w_{1:k-n-1}) \prod_{k=1}^{m} P(w_k|w_{k-n:k-1})$$

$$= P_{context} \prod_{k=1}^{m} P(w_k|w_{k-n:k-1}),$$

where the last term is an n-gram language model and

$$P_{context} = \prod_{k=1}^{m} P(w_{k-n:k-1}|w_{1:k-n-1}) \quad (3)$$

depends on the context only, and is considered to be equal to 1 as an approximation. A frequentist explanation of the approach is that if we consider sufficiently long pieces of the text, they would be unique even in a large corpus, thus all the counts would be 1 and so the conditional probabilities would be (the alternative would be resolving an ambiguity of $\frac{0}{0}$, which can be extrapolated to be equal to 1 as well):

$$P(w_{k-n:k-1}|w_{1:k-n-1}) = \frac{C(w_{1:k-1})}{C(w_{k-n:k-1})} \quad (4)$$

With a Link Grammar, we work with graphs and can actually build a tree of a sentence. In this structure, the context information is beyond the sentence, unlike the n-gram model. Thus, we can start with the root of the tree and use the chain rule along each branch, but we should specifically take the context into consideration, as each conditional probability does depend on the context.

More specifically, let's denote $w_1$ the root of the sentence tree, $w_k$ – a term appearing in the sentence, $w_1/w_k$ – the path from the root to the term $w_k$ (following, for example, Grimmett [7]). Then, we can assume that the probabilities of different branches are independent. What does this assumption/approximation imply requires a separate discussion.

With the notation and assumptions listed above, we can write the probability of the sentence as

$$P(S) = \prod_{k=1}^{m} P(w_k|w_1/w_k^-) \quad (5)$$

Let's take note that in the classical dissertation of Yuret [25] the formula (12) in the proof of Theorem 1 provides only an approximated form of the same probability of a sentence. In our notation, this formula is

$$P(S) = \prod_{k=1}^{m} P(w_k|w_k^-) \quad (6)$$

The difference is small but important. The implicit assumption in (6) is that the conditional probability of a word in a sentence depends on an only one linked word (its pre-decessor). For linear, n-gram models, this would be equivalent to saying that a probability of any n-gram is equal to the probability of its final bigram, which is incorrect from both empirical and mathematical viewpoints. This leads Yuret to an incorrect conclusion that "the entropy of the model is completely determined by the mutual information captured in syntactic relations" of "correlation taken for causation" type.

Further, this leads Yuret to conclude, "The goal of the processor is to find the dependency structure that assigns a given sentence a high probability. In Chapter 3, I showed that the probability of a sentence is determined by the mutual information captured in syntactic relations. Thus, the problem is to find the dependency structure with the highest total mutual information." This approach is ungrounded, as we have seen above, so the approach to building the dependency structure is also incorrect. Unfortunately, many subsequent works have relied on this conclusion (for example, [12] and [23]). This correlates well with the unimpressive results in unsupervised training of graph language models based on bigram approach of Yuret.

However, we must pay tribute to [12] and note that the authors have understood that they work with an assumption: "All systems that we are aware of operate under the assumption that the probability of a dependency structure is the product of the scores of the dependencies (attachments) in that structure." By now, it is clear that the assumption is wrong.

## 5    Acknowledgements

The author is grateful to Anton Kolonin for introduction into Link Grammars and discussions of this work.

## 6    References

1. Altmann, E.G., Cristadoro, G., Degli, M.: On the origin of long-range correlations in texts. PNAS. 109, 11582–11587 (2012). https://doi.org/10.1073/pnas.1117723109.
2. Alvarez-Lacalle, E., Dorow, B., Eckmann, J., Moses, E.: Hierarchical structures induce long-range dynamical correlations in written texts. PNAS. 103, 7956–7961 (2006). https://doi.org/10.1073/pnas.0510673103.
3. Backus, J. W.: The syntax and semantics of the proposed international algebraic language of the Zurich ACM-GAMM Conference. Information Processing: Proceedings of the International Conference on Information Processing, Paris. UNESCO (1959).
4. Brown, T.B., Mann, B., Ryder, N., Subbiah, M., Kaplan, J., Dhariwal, P., Neelakantan, A., Shyam, P., Sastry, G., Askell, A., Agarwal, S., Herbert-Voss, A., Krueger, G., Henighan, T., Child, R., Ramesh, A., Ziegler, D.M., Wu, J., Winter, C., Hesse, C., Chen, M., Sigler, E., Litwin, M., Gray, S., Chess, B., Clark, J., Berner, C., McCandlish, S., Radford, A., Sutskever, I., Amodei, D.: Language models are few-shot learners. In: Advances in Neural Information Processing Systems. pp. 1877–1901 (2020).
5. Chomsky, N.: Three models for the description of language. IRE Transactions on Information Theory, 2(3):113–124 (1956).
6. Delétang, G., Ruoss, A., Grau-Moya, J., Genewein, T., Wenliang, L.K., Catt, E., Hutter, M., Legg, S., Ortega, P.A.: Neural Networks and the Chomsky Hierarchy. (2022).

7. Devlin, J., Chang, M.W., Lee, K., Toutanova, K.: BERT: Pre-training of deep bidirectional transformers for language understanding. In: NAACL HLT 2019 - 2019 Conference of the North American Chapter of the Association for Computational Linguistics: Human Language Technologies - Proceedings of the Conference. pp. 4171–4186 (2019).
8. Grimmett, G.: Probability on Graphs: Random Processes on Graphs and Lattices. Cambridge University Press (2018).
9. Holtzman, A., Buys, J., Du, L., Forbes, M., Choi, Y.: The curious case of neural text degeneration. In: Proceedings of the 2020 International Conference on Learning Representations (2020).
10. Jelinek, F., Lafferty, J.D., Mercer, R.L.: Basic Methods of Probabilistic Context Free Grammars. In: Speech Recognition and Understanding. pp. 345–360 (1992). https://doi.org/10.1007/978-3-642-76626-8_35.
11. Dan Jurafsky and James H. Martin. Speech and Language Processing (3rd ed. draft) (2022)
12. Klein, D., Manning, C.D.: Corpus-Based Induction of Syntactic Structure: Models of Dependency and Constituency. In: Proceedings of the 42nd Annual Meeting on Association for Computational Linguistics. pp. 479–486 (2004). https://doi.org/10.3115/1218955.1219016.
13. Lafferty, J., Sleator, D., Temperley, D.: Grammatical Trigrams: A Probabilistic Model of Link Grammar. AAAI Tech. Rep. FS-92-04. In: Proceedings of the AAAI Conference on Probabilistic Approaches to Natural Language. pp. 89–97 (1992).
14. Lin, H.W., Tegmark, M.: Critical behavior in physics and probabilistic formal languages. Entropy. 19, 1–25 (2017). https://doi.org/10.3390/e19070299.
15. Paskin, M.A.: Grammatical bigrams. In: Proceedings of the 14th International Conference on Neural Information Processing Systems: Natural and Synthetic (NIPS'01). pp. 91–97. MIT Press, Cambridge, MA, USA (2002). https://doi.org/10.7551/mitpress/1120.003.0016
16. Ramesh, V., Kolonin, A.: Natural Language Generation Using Link Grammar for General Conversational Intelligence. 1–17. ArXiv abs/2105.00830 (2021)
17. Ramesh, V., Kolonin, A.: Unsupervised Context-Driven Question Answering Based on Link Grammar. Lect. Notes Comput. Sci. (including Subser. Lect. Notes Artif. Intell. Lect. Notes Bioinformatics). 13154 LNAI, 210–220 (2022). https://doi.org/10.1007/978-3-030-93758-4_22.
18. Sanh, V., Webson, A., Raffel, C., Bach, S.H., Sutawika, L., Alyafeai, Z., Chaffin, A., Stiegler, A., Scao, T. Le, Raja, A., Dey, M., Bari, M.S., Xu, C., Thakker, U., Sharma, S.S., Szczechla, E., Kim, T., Chhablani, G., Nayak, N., Datta, D., Chang, J., Jiang, M.T.-J., Wang, H., Manica, M., Shen, S., Yong, Z.X., Pandey, H., Bawden, R., Wang, T., Neeraj, T., Rozen, J., Sharma, A., Santilli, A., Fevry, T., Fries, J.A., Teehan, R., Bers, T., Biderman, S., Gao, L., Wolf, T., Rush, A.M.: Multitask Prompted Training Enables Zero-Shot Task Generalization. In: ICLR (2022).
19. Shieber, S.M.: Evidence against the context-freeness of natural language. Linguist. Philos. 8, 333–343 (1985). https://doi.org/10.1007/BF00630917.
20. Siegelmann, H.T., Sontag, E.D.: Analog computation via neural networks. Theor. Comput. Sci. 131, 331–360 (1994). https://doi.org/10.1016/0304-3975(94)90178-3.
21. Sleator, D., and Temperley. D.,: Parsing english with a link grammar. In Proc. Third International Workshop on Parsing Technologies, pages 277–292, (1993).
22. Tesni`ere, L.. El´ements de syntaxe structurale´. Klincksieck, Paris, (1959).
23. Vepstas, L., Goertzel, B.: Learning Language from a Large (Unannotated) Corpus. (2014). arXiv:1401.3372
24. https://en.wikipedia.org/wiki/Grammar
25. Yuret, D.: Discovery of linguistic relations using lexical attraction. PhD thesis, MIT, (1998). arXiv preprint cmp-lg/9805009.